\documentclass[runningheads]{llncs}
\usepackage{amsmath,amsfonts,bm}
\usepackage{subfig}
\usepackage{mathtools}
\usepackage{graphicx}
\usepackage{float}
\usepackage{caption}
\usepackage{hyperref} 
\usepackage{cuted}
\usepackage{lipsum, color}

\usepackage{booktabs}
\usepackage{threeparttable}

\usepackage{mathtools}

\DeclarePairedDelimiter\floor{\lfloor}{\rfloor}

\begin{document}

\title{A Study on Graph-Structured Recurrent Neural Networks and  Sparsification with Application to Epidemic Forecasting}

\author{Zhijian Li\inst{1}
\and
Xiyang Luo\inst{2}
\and
Bao Wang\inst{2}
\and
Andrea L. Bertozzi\inst{2}
\and
Jack Xin\inst{1}
}

\authorrunning{Z. Li et al.}
%
\institute{UC Irvine \email{zhijil2@uci.edu, jxin@math.uci.edu}\and
UCLA \email{xylmath@gmail.com, wangbaonj@gmail.com, bertozzi@math.ucla.edu}
}

\maketitle

\begin{abstract}
We study epidemic forecasting on real-world health data by a graph-structured recurrent neural network (GSRNN). We achieve state-of-the-art forecasting accuracy on the benchmark CDC dataset. To improve model efficiency, we sparsify the network weights via a transformed-$\ell_1$
penalty without losing prediction accuracy.  

\end{abstract}

\section{Introduction}
Epidemic forecasting has been studied  for decades \cite{perra2015}. Many statistical and machine learning methods have been successfully used to detect epidemic outbreaks \cite{rev2014}. In previous works, epidemic forecasting is mainly considered as a time-series problem. Time-series methods, such as Auto-Regression (AR), Long Short-term Memory (LSTM) neural networks and their variants have been applied to this problem. 
One of the current directions is to use social media data \cite{volk2017}. In 2008, Google launched Google Flu Trend, a digital service to predict influenza outbreaks using Google search data. The Google algorithm was discontinued due to flaws, however Yang et al. \cite{argo} designed another algorithm ARGO in 2015 also using Google search pattern data. Google Correlate, a collection of time-series data of Google search trends, plays a vital role in this refined regression algorithm. Though ARGO succeeded in accuracy as a time series algorithm, it lacks spatial structure and requires the additional input of external features (e.g., social media data). The infectious and spreading  nature of the epidemics suggests that forecasting is also a spatial problem. Here we study a model to take advantage of the spatial information so that the data from the adjacent regions can introduce regional spatial  features. This way, we minimize  external data input and the accompanying computational cost. Structured recurrent neural network (SRNN) is a model for the spatial-temporal problem first adopted by A. Jain et al. \cite{Jain_16} for  motion forecasting in computer vision. Wang et al. \cite{Wang,Wang:DLCrime,Wang:GSRNN} successfully adapted SRNN to forecast real-time crime activities. Motivated by \cite{Jain_16,Wang,Wang:GSRNN}, we present an SRNN model to forecast epidemic activity levels. We test our model with data provided by the Center for Disease Control (CDC), which collects data from approximately 100 public and 300 private laboratories in the US \cite{cdc}. 
The CDC data \cite{cdc} is a well-established authoritative data set widely used by researchers, which makes it easy for us to compare our model with previous work. CDC provides the influenza data by the geography of Health and Human Services regions (HHS regions). We take the geographic structure of ten HHS regions as our spatial information. The rest of the paper is organized as follows. In section 2, we overview RNN. In sections 3-5, we present a graph-structured RNN model, graph description of spatial correlations, and sparsity promoting penalties. Experimental results and concluding remarks are in sections 6 and 7.

\section{A Short Review of Recurrent Neural Network}
Recurrent Neural Network (RNN) is a neural network designed for sequential data. It is widely used in natural language processing (NLP) and time-series analysis because of its capability of processing sequential information. The idea of RNN comes from unfolding a recursive computation for a chain of states. If we have a chain of states, in which each state depends on the last steps:
$s^n=f(s^{n-1})$, 
for some function $f$.
Then we can unfold this equation to:
$s^n=f(f(...f(s^0)))$.
Suppose we have a sequential data ${x^1,x^2,....,x^n}$, the idea of RNN is to unfold the $x^n=f(x^{n-1},\theta)$ to a computational graph. 
An unfolded RNN is illustrated in Fig.~\ref{RNN:Demo} and given by the recursion:
\begin{figure}[!ht]
    \begin{center}
    \includegraphics[width=6cm,height=6cm,keepaspectratio]{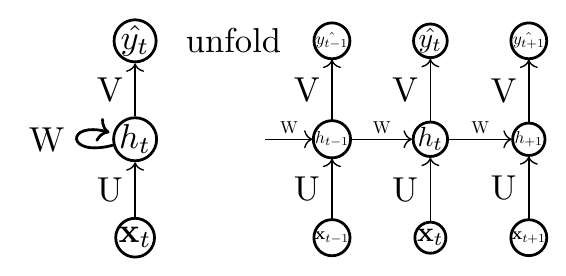}
    \caption{an unfolded recurrent neural network.}
    \label{RNN:Demo}
    \end{center}
\end{figure}
\vskip -0.5in
$$
h_t=\tanh(b+W\, h^{t-1}+U\, x^t), \;\;
y_t=\tanh(V\, h_t+c), 
$$
where tanh is the activation function; $(U,V,W)$ are the weight matrices; $b, c$ are bias vectors. Given $y^t_i$ the true signal at time $t$, the  popular loss function for classification task is the cross-entropy loss, which reads in the binary case:
$\mathcal{L(\theta)}=-\sum_{t}y_t\cdot \ln(\hat{y_t})+(1-y)\ln(1-\hat{y})$.
The mean-square-error loss is widely used for regression problem:
$\mathcal{L(\theta)}=\sum_{t}(y_t-\hat{y_t})^2$.
Then, as in most neural networks, RNN is trained by stochastic gradient descent.
A major issue of RNN is the problem of exploding and vanishing gradients.
Since 
$$\frac{\partial L}{\partial W}=\sum_{t}\frac{\partial L_t(y^t,\hat{y^t})}{\partial W},\;
\frac{\partial L_t}{\partial W}= \sum_{k=0}^{t}\frac{\partial L_n}{\partial \hat{y^t}}\frac{\partial \hat{y^t}}{\partial s_t}  \frac{\partial s_t}{\partial s_k}\frac{\partial s_k}{\partial W},\;
s_t=\tanh(Ux^t+Wh^{t-1}),
$$

$${\rm then}\; \;\;\;
\frac{\partial L_{t}}{\partial W}=\frac{\partial L_t}{\partial \hat{y^t}} \frac{\partial \hat{y^t}}{\partial s_t} \big ( \prod_{j=k+1}^{t} \frac{\partial s_j}{\partial s_{j-1} }\big ) \frac{\partial s_k}{\partial W}.
$$

It is well-known \cite{Pascanu:2013:DTR:3042817.3043083} that 
$$||\frac{\partial \hat{y^t} }{\partial s_t} \big (\prod_{i=k}^{t-1} \frac{\partial s_{i+1}}{\partial s_{i}}\big ) ||\leq \eta^{t-k}||\frac{\partial s_t}{\partial \hat{y^t}}||
$$
where $\eta < 1$ under the assumption of no bias is used and the spectral norm of $W$ being less than 1. We see that the gradient vanishes exponentially fast in large $t$. Hence, the RNN is learning less and less as time goes by. 
LSTM \cite{LSTM}, is a special kind of RNN that resolves this problem.

$$
\Bigg(\begin{tabular}{c}
$f_t$ \\$i_t$\\$o^t$\\ $\tilde{C}_t$
\end{tabular}\Bigg)
=
\Bigg(\begin{tabular}{c}
$\sigma(W[h_{t-1,x_t}])$ \\ $\sigma(W[h_{t-1,x_t}])$ \\ $\sigma(W[h_{t-1,x_t}])$ \\ $\tanh(W[h_{t-1,x_t}])$
\end{tabular}\Bigg)
+
\Bigg(\begin{tabular}{c}
 $b_f$ \\$b_i$\\$b_o$\\$b_C$
\end{tabular}\Bigg)
$$
$$C_t=f_t*C_{t-1}+i_t*\tilde{C}_t, \;\; h_t=o_t*\tanh(C_t).$$
Since one does not directly apply the same recurrent function to $h_t$ every time step in the gradient flow, there is no intrinsic factor $\eta$ in  $\frac{\partial L_t}{\partial W}$. 
This way the gradient has much less chance to vanish as time goes by. In our model, we use LSTM for all RNNs.

\section{Graph-Strutured RNN model}
\begin{figure}[!ht]
    \begin{center}
    \includegraphics[width=0.75\textwidth]{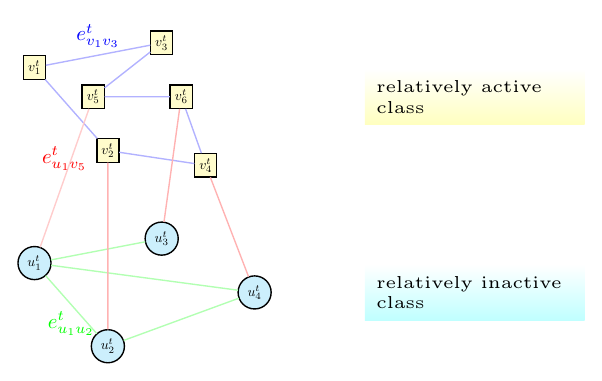}
    \caption{red edges are of type H-L, green edges are of type L-L, and blue edges are of type H-H.}
    \label{figure2}
    \end{center}
\end{figure}

Similar to previous work of structured RNN, we partition the nodes into different classes, and for each class we join the nodes in the class. We compare the level of activity of nodes by summing up the data of each node. Then, we partition the nodes based on their activity level, from the class that has highest activity level to the lowest one. After some experiments, we find that SRNN works the best when we have two classes (see Fig. \ref{figure2}). We denote the class with relatively high activity level H, and the other class L. After some experiments, we classify the nodes based on following criteria: Let $G$ be a weighted graph with nodes indexed by $Z = \{1, \dots, N\}$, and edge weights $w_{ij} \geq 0$. Let $g : Z \mapsto \{1, \dots, C\}$ be the function that assigns each node to its corresponding group, and assume for simplicity that $g$ is a surjection. Let us define: 
$$|v|=\text{sum of the historical activity level of node }v,$$
$$M=\operatorname*{arg\,max}_v |v|\text{,  }\;\; m=\operatorname*{arg\,min}_v |v|,\;\;
g(v)=\floor{\frac{|v|-m}{M-m+10^{-6}}}.$$
In our model, nodes with label 0 are in the relatively inactive class, nodes with label 1 or higher belong to another class, the relatively active class. 

  
We define an RNN $E_{i,j}$ for each connected edge $w_{ij} \neq 0$. We denote $E_{i,j}$ as the \emph{edge} RNN since it models the pairwise interaction between two connected nodes. We enforce weight sharing among two edge RNNs, $\text{RNN}_{E_{i', j'}}$ and $\text{RNN}_{E_{i,j}}$, if $g(i) = g(i')$, $g(j) = g(j')$, i.e., if the class assignments of the two node pairs are the same. 
Similarly, we define an RNN $N_i$, for each node in $Z$, which we denote as a \emph{node} RNN, and apply weight sharing if $g(i') = g(i)$. Even though the RNNs share weights, there state vector are still different, and thus we denote them with distinct indices. 

Let $\{v^t_i, i\in 1\dots N\}$ be the set of node features at time $t$. The GSRNN makes a prediction at node $i$, time $t$ by first feeding neighboring features to its respective edge RNN, and then feeding the averaged output along with the node features to the respective node RNN. Namely, 
\begin{equation}
f^t_i  =  \sum_{j} w_{ij} \text{RNN}_{E_{i, j}} (v^t_i, \alpha_{ij}v^t_j), \;\;\;
\hat{y}^t_i  =  {\rm RNN}_{N_i} (v^t_i, f^t_i). 
\end{equation}
Let $y^t_i$ be the true signal at time $t$. We use the mean square loss function below:
\begin{eqnarray}
L^t(\Theta) = \frac{1}{N} \sum_{i} \;  (\hat{y}^t_i - y^t_i)^2. 
\end{eqnarray}
We back-propagate through time (BPTT), a standard method for training RNNs, with the understanding that the weights for edge RNNs and node RNNs are shared according to the description above.

In our model of $C=2$, we have three types of edges, H-H, L-L, and H-L. The H-H is the type of edge between two nodes in class H, L-L is the type of edge between two nodes of class L, and H-L is the type of edge edge a node of class H and a node of class L. Each type of edge features will be fed into a different RNN. We normalize our edge weight by maximum degree. Each edge has weight $\alpha_{ij}=w_{ij}=\frac{1}{M_e}$, $\forall \, i$ and $j$, where $M_e$ is the maximum degree over the ten nodes. We use a look-back window of two to generate training data for RNN: the node feature of $v^t$ contains the information of node $v$ at $t-1$ and $t-2$. Then, the edge features of a node $v \in H$ with the edges $E_v$ are:
$$
    e_{v,H}^{t}=\left [\frac{v_{1}^t}{M_e},\frac{v_{2}^t}{M_e}\cdots \right ]
$$
for all $v_{i}\in H$ such that $(v,v_i)\in E_v$,
$$e_{v,L}^t=\left [\frac{u_{1}^t}{M_e},\frac{u_{2}^t}{M_e}\cdots \right ]$$
for all $u_{i}\in L$ such that $(v,u_i) \in E_v$. We feed $e_{v,H}^t$ and $e_{v,L}^t$ into the corresponding edgeRNNs:
$$f^{t}=\frac{1}{M_e}\text{edgeRNN}_{H-L}(v^t,e_{v,L}^t),\;\;
h_{v}^t=\frac{1}{M_e}\text{edgeRNN}_{H-H}(v^t,e_{v,H}^{t}).$$
Each edge RNN will jointly train all the nodes that have an edge belong to its type:
$$\operatorname*{arg\,min}_\theta \mathcal{L}_{H-L}(\Theta)=\frac{1}{|N_w|}\sum_{w\in N_w}\sum_{t}({y_w}^{t}-{\hat{y}_w}^{t})^2$$
where $N_w=\{w\in H\cup L | \ \ E_w \text{ contains an element of type L-H}\}$.
$$\operatorname*{arg\,min}_\theta \mathcal{L}_{H-H}(\Theta)=\frac{1}{|N_v|}\sum_{v\in N_v}\sum_{t}({y_v}^{t}-{\hat{y}_v}^{t})^2$$
where $N_v=\{v\in H| \ \ E_v\text{ contains an element of type H-H}\}$.\\
Finally, we have a node RNN that jointly trains all the nodes in this class:
$$\operatorname*{arg\,min}_\theta \mathcal{L}_{H}(\Theta)=\frac{1}{|H|}\sum_{v\in H}\sum_{t}({y_v}^{t}-{\hat{y}_v}^{t})^2, \;\; \forall v \in H.$$
We feed the outputs of two edge RNNs, together the node feature of $v$ itself into $\text{nodeRNN}_{H}$:
$$v^{t+1}=\text{nodeRNN}_{H}(v^t, f^t, h^t).$$
\begin{figure}[!ht]
    \begin{center}
    \includegraphics[width=0.65\textwidth]{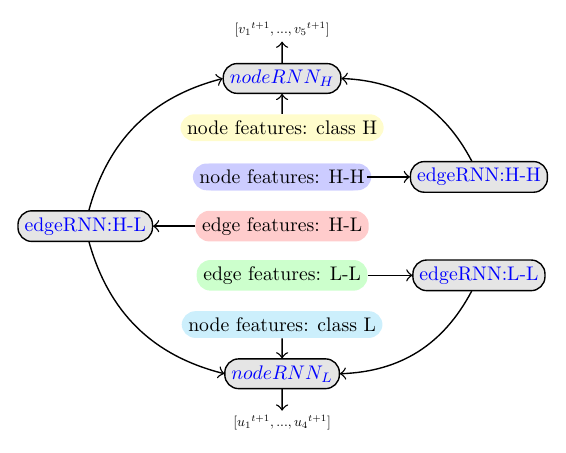}
    \caption{Edge features of the same type are jointly trained by one edge RNN. Nodes from the same class are jointly trained by one node RNN.}
    \label{figure3}
    \end{center}
\end{figure}

\section{Graph Description of Spatial Correlation}
\begin{figure}%
    \subfloat[ HHS regions on map]{{\includegraphics[width=5.5cm,height=4.5cm]{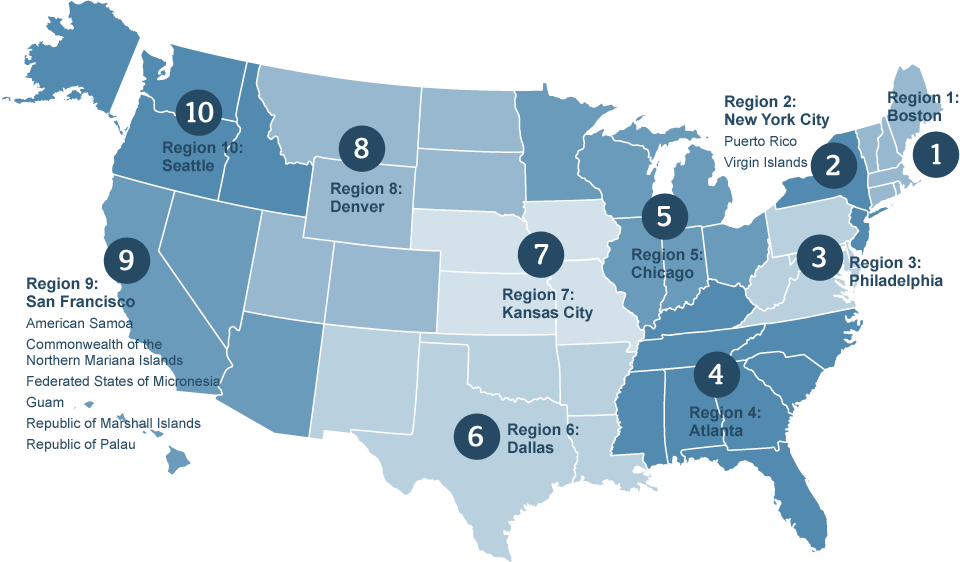} }}%
    \qquad
    \subfloat[Graph Structure]{{\includegraphics[width=5.5cm,height=4.5cm]{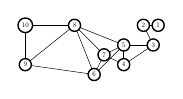} }}%
    \caption{HHS Graph}%
    \label{fig:example}%
\end{figure}

The graph is a flexible representation for irregular geographical shapes which is especially useful for many spatio-temporal forecasting problems. In this work, we use a weighted directed graph for space description where each node corresponds to a state. There are multiple ways to infer the connectivity and weights of this weighted directed graph. In the previous work \cite{Wang:GSRNN}, Wang et al. utilized a multivariate Hawkes process to infer such a graph for crime and traffic forecasting, where the connectivity and weight indicate the mutual influence between the source and the sink nodes. Alternatively, one can opt for space closeness and connect the closest few nodes on the graph, and the weight is proportional to the historical moving average activity levels of the source node. In this work, we employ the second strategy, where we regard two nodes as connected if the corresponding two states are geographically adjacent to each other. The graph in this work is demonstrated in Fig.~\ref{fig:example}. We will explore the first strategy in the future.

\section{Sparsity Promoting Penalties}
The convex sparsity promoting penalty is 
the $\ell_1$ norm. In this study, we also  employ a Lipschitz continuous non-convex penalty, the so-called transformed-$\ell_1$.
\begin{definition}\label{TL1Definition}
The transformed $\ell_1$ (T$\ell_1$) penalty function on $x =(x_1,\cdots, x_d) \in \mathbb{R}^d$ is
\begin{equation} P_a(x) := \sum_{i=1}^d \rho_a(x_i),\;\;\; \rho_a(x_i) = \frac{(a+1)|x_i|}{a+|x_i|},\;\; \; 
{\rm parameter}\;\; a \in (0,+\infty).\label{Tl1}\end{equation}
\end{definition}

Since $ \lim_{a \to 0^+} \rho_a(x_i) = 1_{\{x_i \ne 0\}}$, $\lim_{a \to +\infty} \rho_a(x_i) = |x_i|$, $\forall i$, the T$\ell_1$ penalty interpolates $\ell_1$ and $\ell_0$. For its sparsification in compressed sensing and other applications, see  \cite{TL1} and references therein. To sparsify weights in GSRNN training via  $\ell_1$ and T$\ell_1$,  
we add them to the loss function of GSRNN with a multiplicative penalty parameter $\alpha > 0$, and call stochastic gradient descent optimizer on Tensorflow. Though a relaxed splitting method \cite{RVSM} can enforce sparsity much faster, we shall leave this as part of future work on $\ell_0$ penalty. 
 
\medskip

\section{Experimental Results}

Among the previous works on influenza  forecasting, ARGO \cite{argo} is the current state-of-the-art prediction model for the entire U.S. influenza activity. 
To compare with previous works conveniently, we use the CDC data from 2013 to 2015 as our test data. The accuracy is measured in:
    RMSE=$\sqrt{\frac{1}{n}\sum_{i=1}^{n}{(y_i-\hat{y_i})^2}}$.
\medskip

\begin{figure}[!h]
\centering
\begin{tabular}{cc}
\includegraphics[scale=0.25]{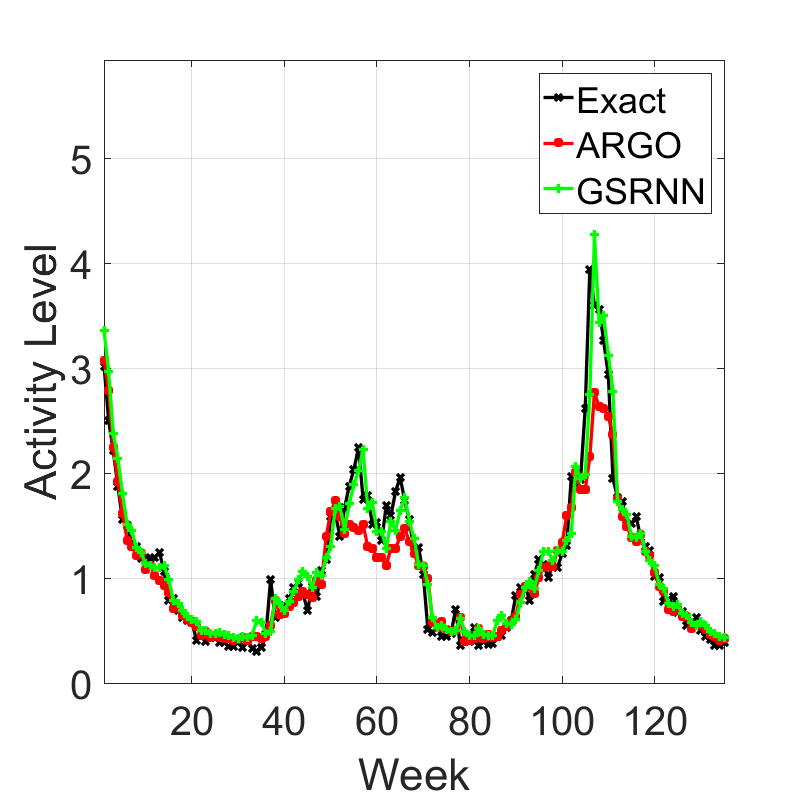}&
\includegraphics[scale=0.25]{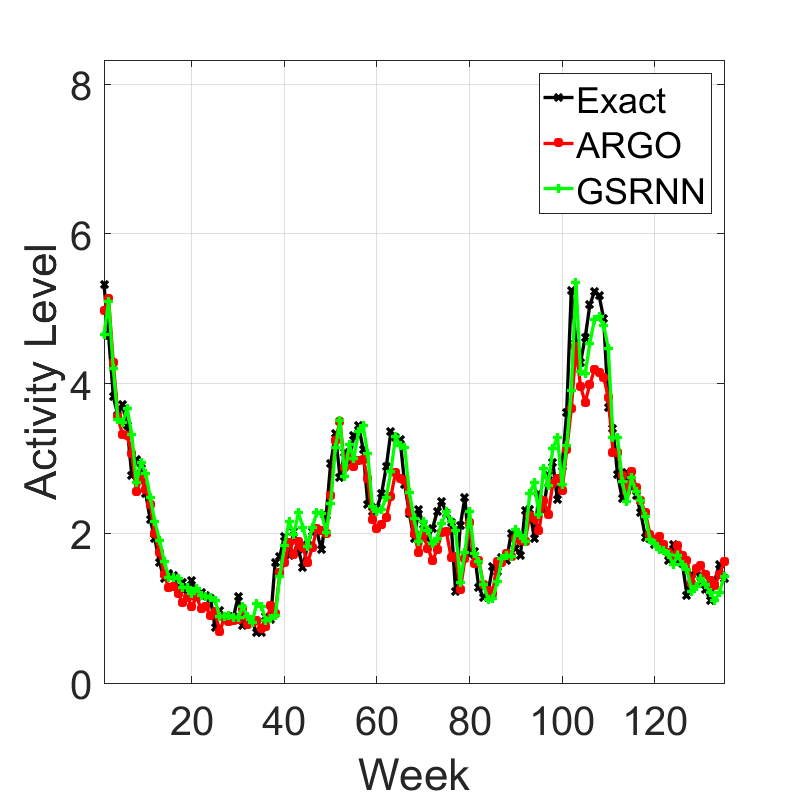}\\
(HHS Region 1) & (HHS Region 2) \\
\includegraphics[scale=0.25]{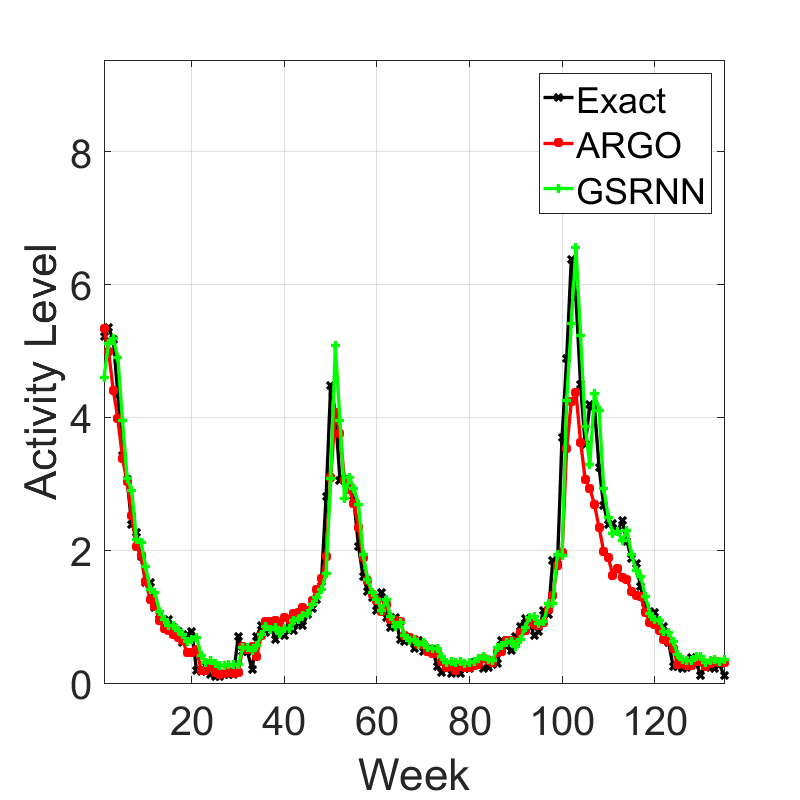}&
\includegraphics[scale=0.25]{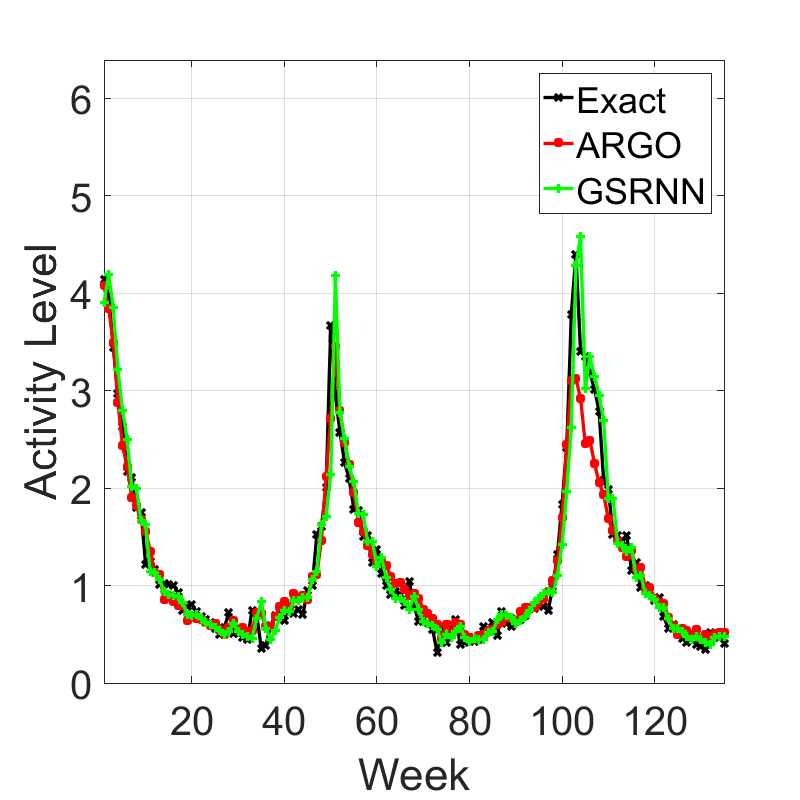}\\
(HHS Region 7) & (HHS Region 8)\\
\end{tabular}
\caption{The exact and predicted flu activity levels by GSRNN and ARGO.}
\label{fig:Comparison}
\end{figure}

We use a single layer LSTM with 40 hidden units for edge RNNs, and a three-layer multilayer LSTM with hidden units $[10,40,10]$ for node RNNs. We use the Adam optimizer to train GSRNN. The RMSE of the forecasting from 2013/1/19 to 2015/8/15, 135 weeks in total, is shown in Table.~\ref{Table1:RMSE:Comparison}. We outperform LSTM and Autoregressive Model of order 3 (AR(3)) in all nodes, and ARGO in 8 nodes, see Fig.~\ref{fig:Comparison} for activity plots in each region. It is easy to see that in regions 1, 2, 7 and 8, there are some under-predictions, while GSRNN's prediction is almost identical to the ground-truth. The general form of an AR(p) model for time-series data is $$X_t=\mu+\sum_{i=1}^p \; \phi_i \, X_{t-i}+\epsilon,$$
where $\vec{\phi}=(\phi_1,...,\phi_p)$ is computed 
through the backshift operator. ARGO\cite{argo}, as a refined autoregressive model, models the flu activity level as:
$$\hat{y_t}=\mu_y+\sum_{j=1}^{52}\; \alpha_j \, y_{t-j}+\sum_{i=1}^{100}\; \beta_i \, X_{i,t}+\epsilon_t, \;\; \; 
[\mu_y,\,\vec{\alpha},\, \vec{\beta}] := \operatorname*{arg\,min}_{\mu_y,\,\vec{\alpha},\, \vec{\beta}}\sum_{t}\; (y_t-\hat{y_t})^2,$$ 
$\epsilon_t$ being i.i.d Gaussian noise, $X_{i,t}$ the log-transformed Google search frequency of term $i$ at time $t$. \\

We observe that ARGO has inconsistent performance over nodes. We believe this is because the external feature of ARGO, the Google search pattern data, does not offer useful information, since the national search pattern does not necessarily apply to a certain HHS region. Meanwhile, we also have much less computational cost than ARGO, which takes in top 100 search terms related to influenza as well as their historical activity levels, with a look-back window length of 52 weeks. During the time for ARGO to compute one node, our model finishes all the ten nodes. \\

We sparsify the network through $\ell_1$ and T$\ell_1$ (eq. (\ref{Tl1}) using $a=1$ and penalty parameter $\alpha=10^{-8}$ during training.
Post training, we hard threshold small network weights to 0  
at threshold $10^{-3}$, and find that high sparsity under T$\ell_1$ regularization is achieved while maintaining the accuracy at the same level, see Table \ref{table3} and Table \ref{table4}. 
Hard-thresholding improves the predictions for some nodes but not all of them, however it reduces the inference latency and is thus beneficial for the overall algorithm.


\begin{table*}[h]
\centering
\fontsize{10.0}{10}\selectfont
\begin{threeparttable}
\caption{The RMSE between the predicted and ground-truth activity levels by different methods over 10 different states. }\label{Table1:RMSE:Comparison}
\begin{tabular}{ccccccccccc}
\toprule
Node     & 1  & 2   & 3  & 4 & 5 & 6 & 7 & 8  & 9 & 10\cr
\midrule
AR(3)       &0.242 &0.383 &0.481 &0.415 &0.345 &0.797 &0.401 &0.305 &0.356 &0.317 \cr
ARGO        &0.281 &0.379 &0.397 &0.335 &{\bf 0.285} &0.673 &0.449 &{\bf 0.244} &0.356 &0.310 \cr
LSTM        &0.271 &0.364 &0.487 &0.349 &0.328 &0.751 &0.421 &0.333 &0.335 &0.310 \cr
GSRNN       &{\bf 0.223} &{\bf 0.354} &{\bf 0.374} &{\bf 0.320} &0.289 &{\bf 0.664} &{\bf 0.361} &0.275 &{\bf 0.284} &{\bf 0.303} \cr
\bottomrule
\end{tabular}
\end{threeparttable}
\end{table*}

\medskip

\begin{table*}[h]
\centering
\fontsize{10.0}{10}\selectfont
\begin{threeparttable}
\caption{Percentages of weights $<10^{-3}$ in absolute value in 
GSRNN w/ and w/o $\ell_1$, T$\ell_1$ penalties. 
}\label{table3}
\begin{tabular}{cccccc}
\toprule
Penalty     & 1  & 2   & 3  & 4 & 5 \cr
\midrule
$\alpha=0$                   &51.2\% &47.8\% &50.3\% &50.6\% &49.9\%  \cr
$l_1(\alpha=5\cdot 10^{-8})$ &67.7\% &51.8\% &57.7\% &60.7\% &61.2\%  \cr
$TL1(\alpha=5\cdot 10^{-8})$ &{\bf 82.3}\% &{\bf 58.9}\% &{\bf 71.9}\% &{\bf 64.2}\%& {\bf 71.1}\%  \cr
\bottomrule
\end{tabular}
\end{threeparttable}
\end{table*}

\begin{table*}[h]
\centering
\fontsize{10.0}{10}\selectfont
\begin{threeparttable}
\caption{Node-wise RMSE of GSRNNs via post-training hard thresholding at threshold $10^{-3}$.}\label{table4}
\begin{tabular}{ccccccccccc}
\toprule
Node     & 1  & 2   & 3  & 4 & 5 & 6 & 7 & 8  & 9 & 10\cr
\midrule
$\alpha=0$                   &0.230 &0.351 &0.390 &0.334 &0.314 &0.676 &0.380 &0.297 &0.287 &0.316 \cr
$l_1(\alpha=5\cdot 10^{-8})$ &0.234 &0.351 &0.388 &0.327 &0.306 &0.685 &0.363 &0.290 &0.281 &0.296 \cr
$TL1(\alpha=5\cdot 10^{-8})$ &0.225 &0.363 &0.379 &0.328 &0.296 &0.690 &0.365 &0.272 &0.311 &0.305 \cr
\bottomrule
\end{tabular}
\end{threeparttable}
\end{table*}

\section{Concluding Remarks}
We studied epidemic forecasting based on a graph-structured RNN model to take into account geo-spatial information. 
We also sparsified the model and reduced 70\% of the network weights to zero while maintaining the same level of prediction accuracy. In future work, we plan to explore wider neighborhood interactions and more powerful sparsification methods. Furthermore, training RNNs with the recently developed Laplacian smoothing gradient descent \cite{LSGD:2018} is worth exploring.


\section*{Acknowledgments}
This material is based on research sponsored by the Air Force Research Laboratory and DARPA under agreement number FA8750-18-2-0066; the U.S. Department of Energy, Office of Science, DOE-SC0013838; the  National Science Foundation DMS-1554564 (STROBE), DMS-1737770, DMS-1522383, IIS-1632935. 
The authors thank Profs. 
M. Hyman, and J. Lega for helpful discussions.  

\thispagestyle{empty}







\end{document}